\pdfoutput = 1
\documentclass[8pt]{article}

\usepackage{times}
\usepackage[comma,authoryear]{natbib}
\usepackage[english]{babel}
\usepackage{blindtext}
\usepackage{graphicx}
\usepackage{amsmath, amsthm, amssymb, bbm, bm}
\usepackage{enumerate}
\usepackage{float}      
\usepackage{subcaption}  
\usepackage{wrapfig}
\usepackage[margin=0cm]{caption}
\usepackage[titletoc, toc]{appendix}
\usepackage{tabularx}

\usepackage{multirow}
\usepackage{makecell}

\usepackage[x11names, usenames, dvipsnames, svgnames, table]{xcolor}
\definecolor{firebrick}{rgb}{.698,.133,.133}

\usepackage[utf8]{inputenc} 
\usepackage[T1]{fontenc}    
\usepackage{url}            
\usepackage{booktabs, colortbl}       
\usepackage{amsfonts}       
\usepackage{nicefrac}       
\usepackage{microtype}      

\usepackage{csquotes}
\usepackage{latexsym}

\usepackage[boxruled, vlined, linesnumbered]{algorithm2e}
\SetAlFnt{\small}
\SetAlCapFnt{\small}
\SetAlCapNameFnt{\small}
\usepackage{algorithmic}
\algsetup{linenosize=\tiny}

\let\oldnl\nl
\newcommand{\nonl}{\renewcommand{\nl}{\let\nl\oldnl}}

\usepackage{paralist}

\usepackage{xspace}
\usepackage{soul}
\usepackage{dsfont}
\usepackage{stmaryrd}
\usepackage[textwidth=15mm]{todonotes}
\usepackage{dirtytalk}
\usepackage{pbox}
\usepackage{paralist}
\usepackage{cprotect}

\usepackage{verbatim}
\usepackage{textcomp}
\usepackage[normalem]{ulem}

\usepackage{mathtools}
\usepackage{etextools}
\usepackage[inline]{enumitem}

\usepackage[colorlinks=true,allcolors=firebrick,bookmarks=false]{hyperref}

\newcommand{\R}{\mathbb{R} \xspace}

\newcommand{\abs}[1]{\ensuremath \left| #1 \right|}

\DeclareMathOperator*{\argmax}{argmax}

\theoremstyle{definition}

\newcommand{\E}{\mathop{\mathbb{E} \xspace}}

\providecommand{\ind}{{\bf 1}}

\DeclarePairedDelimiterX{\divx}[2]{(}{)}{%
  #1\;\delimsize\|\;#2%
}

\newcommand{\before}{\prec \xspace}
\DeclareMathOperator{\dom}{\textbf{dom}}  
\DeclareMathOperator{\ran}{\textbf{ran}}  

\usepackage{style}

\title{Non-parametric Uni-modality Constraints \\ for Deep Ordinal Classification
\thanks{Code:~\href{https://github.com/sbelharbi/Deep-Ordinal-Classification-with-Inequality-Constraints}{https://github.com/sbelharbi/Deep-Ordinal-Classification-with-Inequality-Constraints}}}

\renewcommand\footnotemark{}

\renewcommand\footnotemark{}

\author{Soufiane Belharbi$^{1}$, Ismail Ben Ayed$^{1}$, Luke McCaffrey$^{2}$, Eric Granger$^{1}$\\[0.03in]
$^{1}$ LIVIA, \'Ecole de technologie sup\'erieure, Universit\'e du Qu\'ebec, Montreal, Canada\\
$^{2}$ Rosalind and Morris Goodman Cancer Research Centre,   Dept. of Oncology,\\ \hspace{1.9mm} McGill University, Montreal, Canada\\ [0.05in]
{\footnotesize
 Email:\ \href{mailto:soufiane.belharbi.1@ens.etsmtl.ca}{soufiane.belharbi.1@ens.etsmtl.ca},
\href{mailto:ismail.benayed@etsmtl.ca}{ismail.benayed@etsmtl.ca},
\href{mailto:luke.mccaffrey@mcgill.ca}{luke.mccaffrey@mcgill.ca}
}\\[0.03in]
{\footnotesize
\hspace{8.5mm} \href{mailto:eric.granger@etsmtl.ca}{eric.granger@etsmtl.ca}
}}

\newcommand{\ignore}[1]{}

\begin{document}

\maketitle

\begin{abstract}
We propose a new constrained-optimization formulation for deep ordinal classification, in which uni-modality of the label distribution is enforced implicitly via a set of inequality constraints over all the pairs of adjacent labels. Based on $(c-1)$ constraints for $c$ labels, our model is {\em non-parametric} and, therefore, more flexible than the existing deep ordinal classification techniques. Unlike these, it does not restrict the learned representation to a single and specific parametric model (or penalty) imposed over all the labels. Therefore, it enables the training to explore larger solution spaces, while removing the need for {\em ad hoc} choices, and scaling up to large numbers of labels. Our formulation can be employed in conjunction with any standard classification loss and deep architecture.
To address this challenging optimization problem, we solve a sequence of unconstrained losses based on a powerful extension of the log-barrier method. This effectively handles competing constraints and accommodates standard SGD for deep networks, while avoiding computationally expensive Lagrangian dual steps and substantially  outperforming penalty methods. Furthermore, we propose a new Sides Order Index (SOI) performance metric for ordinal classification, as a proxy to measure distribution uni-modality. We report comprehensive set of evaluations and comparisons with state-of-the-art methods on benchmark public datasets for several ordinal classification tasks, showing the merits of our approach in terms of label consistency, classification accuracy and scalability. Importantly, enforcing label consistency with our model does not incur higher classification errors, unlike many existing ordinal classification methods.
\end{abstract}

\section{Introduction}
\label{sec:introduction}

Different research has suggested using a set of surrogate losses for OC training \citep{pedregosa2017consistency,rennie2005loss}. However, cross-entropy and/or mean squared error remain a primary choice for deep models due to their differentiability, simplicity, and capability to deal with many classes. Such standard classification losses
, e.g., cross-entropy, do not impose any prior on the structure of the labels. Designed to penalize the error between predicted and true labels for each data sample, they do not account for the semantic relationships that might exist between the labels. However, in a wide range of classification tasks, the set of labels exhibits a natural structure, for instance, in the form of a specific order. A typical example is classifying biopsy samples, with the labels encoding cancer-aggressiveness levels (or grades), which are ordered. Ordinal classification (OC) attempts to leverage such natural order of the labels, and is useful in a breadth of applications, such as movies rating \citep{crammer2002pranking,Koren2011}, market bonds rating \citep{moody1994architecture}, age estimation \citep{liu2017ordinal,pan2018mean,zhu2019ordinal}, emotion estimation \citep{jia2019facial,XiongLiuZhongFu2019,zhou2015emotion}, cancer grading \citep{gentry2015penalized}, diabetic retinopathy grading \citep{beckham2017unimodal}, photographs dating \citep{palermo2012datingdataset}, among many others.

\begin{figure}[h!]
  \centering
  \includegraphics[scale=0.5]{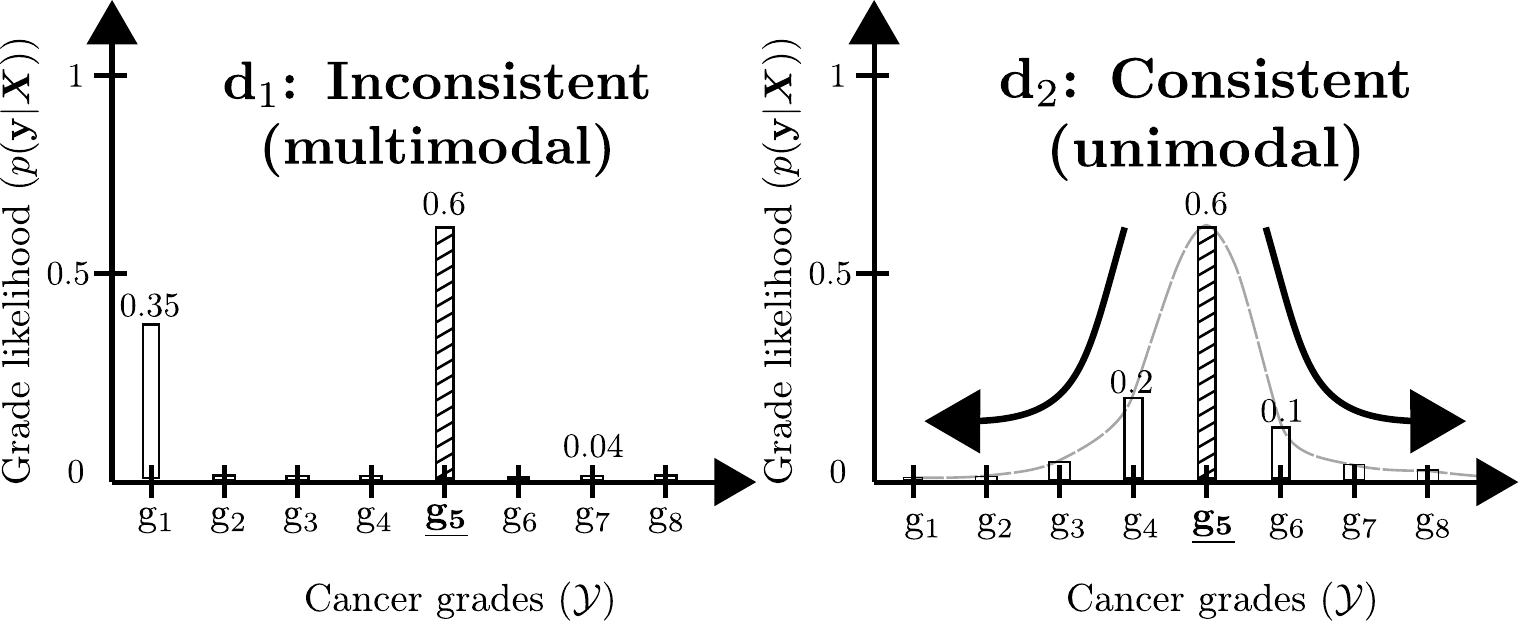}
  \caption{An example of two different posterior distributions of the same input sample with the same probability of the predicted label ${\text{g}_5}$ and the same cross-entropy loss: ${\text{d}_1}$ corresponds to paradoxical predictions, ranking grade ${\text{g}_1}$ right after predicted grade ${\text{g}_5}$, despite their significant semantic difference, whereas ${\text{d}_1}$ corresponds to consistent (ordered) predictions.}
  \label{fig:fig-0}
\end{figure}

Fig.\ref{fig:fig-0} depicts a typical OC task, which consists of grading cancer (8 classes, for illustration). It shows two different posterior distributions (PDs) of class predictions for the same input. While the two PDs have exactly the same cross-entropy loss, ${\text{d}_2}$ corresponds to a {\em uni-modal} distribution of the posteriors as it
concentrates its probability mass around predicted label ${\text{g}_5}$ and, as we move away from the latter, it decreases {\em monotonically}. In contrast, ${\text{d}_{1}}$ distributes its mass over labels that are far away from each other, yielding inconsistent class predictions. Therefore, ${\text{d}_1}$ corresponds to predictions that are semantically paradoxical. The model gave the second highest posterior probability to grade ${\text{g}_1}$, ranking it right after predicted grade ${\text{g}_5}$, despite the significant semantic difference between the two grades. In practice, such inconsistent distributions may raise serious interpretability issues, and impede the deployment of the models, more so when important actions are associated with the predictions. As an example, consider ${\text{d}_1}$ with the following ordered actions for the top-3 labels $\{\text{mild-chemotherapy}, \text{do-nothing}, \allowbreak \text{immediate-surgery-with-intensive-chemotherapy}\}$. This order of top-3 actions is confusing, and may not be considered due to its discordance.

Several recent deep learning works addressed ordinal classification by imposing a uni-modality prior on the predicted posterior distributions following some {\em parametric} model. This is often done by enforcing uni-modality either on the label distributions \citep{beckham2016simple,cheng2008neural,gao2017deep,geng2016label,geng2013facial,huo2016deep,pan2018mean} or on the network outputs
\citep{beckham2017unimodal,da2005classification}. While both type of approaches yield consistent predictions, they
constrain  the  output  distribution  to  have  a  specific form, following the choice of a parametric uni-modal prior, e.g., Poisson \citep{beckham2017unimodal} or Gaussian \citep{gao2017deep,geng2016label,geng2013facial,huo2016deep}, imposed as a {\em single} penalty on all the labels . Therefore, in general, they require
several task-dependent choices, including tuning carefully the hyper-parameters that control the form (or shape) of the parametric model and complex network-architecture design. They might also lead to models that do not scale well for large numbers of labels, as is the case of \citep{beckham2017unimodal}\footnote{The experiments in \citep{beckham2017unimodal} were limited to 8 labels}, which limits their applicability.
We argue that restricting the posterior distribution to a single parametric model of a specific form is not necessary for ensuring uni-modality and order consistency, and propose a non-parametric,  constrained-optimization solution for deep ordinal classification.

{\bf Contributions:} We propose a novel constrained-optimization formulation for deep ordinal classification. Our model enforces uni-modality and label-order consistency via a set of inequality constraints over all pairs of adjacent labels, which can be imposed on any standard classification loss and integrated with any deep architecture. Based on $(c-1)$ constraints for $c$ labels, our model is {\em non-parametric} and, therefore, more flexible than the existing deep ordinal classification techniques. Unlike these, it does not restrict the learned representation to a single and specific parametric model (or penalty) imposed on all the labels. Therefore, it enables the training to explore larger spaces of solutions, while removing the need for {\em ad hoc} choices and scaling up to large numbers of labels.

To tackle the ensuing challenging optimization problem, we solve a sequence of unconstrained losses based on a powerful extension of the log-barrier method.
This handles effectively competing constraints and accommodates standard SGD for deep networks, while avoiding computationally expensive Lagrangian dual steps and outperforming substantially penalty methods. Furthermore, we introduce a new performance metric for OC, as a proxy to measure distribution uni-modality, referred to as the Sides Order Index (SOI). We report comprehensive evaluations and comparisons to state-of-the-art methods on benchmark public datasets for several OC tasks (breast cancer grading, age estimation, and historical color image dating). The results indicate that our approach outperforms substantially several state-of-the-art ordinal classification methods in terms of label consistency, while scaling up to large numbers of labels. Importantly, enforcing label consistency with our model does not incur higher classification errors, unlike many existing ordinal classification methods.

\section{Related work}
\label{sec:related-work}

Several recent work converted the hard target label into a prior distribution \citep{gao2017deep,geng2016label,geng2013facial,huo2016deep}.
One way to impose such a prior distribution over the labels is, for instance, to optimize a divergence loss, such as Kullback–Leibler (KL), for training the network \citep{geng2016label}. A typical choice in the literature is to use a parametric uni-modal Gaussian to model label distribution, with the mean of the Gaussian encoding the true label, while the variance is set through validation or through prior knowledge. The main motivation behind this class of label-distribution methods is to deal with the ambiguity and uncertainty of discrete labels, in tasks such as age estimation, head pose estimation, and semantic segmentation. Using prior distributions over the labels is also related to the well-known label-smoothing regularization for improving the training of deep networks \citep{szegedy2016rethinking}. Such a regularization perturbs the hard label with a uniform distribution, embedding uncertainty in the ground truth. In \citep{cheng2008neural}, the prior order of the labels is encoded using a step function. Instead of a standard one-hot encoding, a binary-vector encoding the labels is used as a target for network training. Another class of methods in literature penalize directly the deviations of the softmax predictions of the network from a uni-modal Gaussian, which is constrained to have the same mean as the true label and minimal variance \citep{beckham2016simple,pan2018mean}. These methods do not impose specific prior knowledge on the variance, but attempt to push it towards zero. In \citep{beckham2017unimodal,da2005classification} specific parametric distributions are directly encoded within the network output, e.g., binomial or Poisson distributions. The network output is a single scalar used to model the output distribution and to infer the probability of each label. Other methods \citep{liu2018constrained,xia2007recursive} seek to reinforce the order between samples of adjacent labels within the feature space, not in the output space. However, such methods require considerable changes to the network architecture \citep{liu2018constrained}.

 While the above mentioned techniques enforce the prior order and uni-modality of the output distributions, they have several shortcomings. The techniques in \citep{gao2017deep,geng2016label,geng2013facial,huo2016deep} constrain the output distribution to have a specific form (or shape), following the choice of a parametric uni-modal model, e.g., Gaussian. This requires {\em ad hoc} (manual) setting of model parameters, e.g., the variance of a Gaussian, which might have a direct but unclear impact on the results. The models in \citep{beckham2016simple,pan2018mean} learn such variance parameter by pushing it towards zero, yielding a sharp Gaussian that approaches a Dirac function. Choosing a sharp or flat Gaussian has a direct impact on the labels and their order, but it is not clear how to make such a choice. In general, the choices that one has to make as to the form of the parametric model are task-dependent. Directly encoding a specific parametric distribution within the network output, as in \citep{beckham2017unimodal}, also requires several choices, including complex network architecture design and setting {\em ad hoc} parameters. In particular, the performance of the method in \citep{beckham2017unimodal} seems to depend strongly on a hyper-parameter that controls the variance of the distribution. Such a hyper-parameter should be set empirically with some care since its value changes the distribution shape from uniform to Gaussian.  The method in \citep{beckham2017unimodal} was evaluated on only two ordinal datasets that have very few classes (5 and 8). The authors of \citep{beckham2017unimodal} stated that the method does not scale well to large numbers of labels, in particular, with the Poisson distribution, due to the nature of the latter. With this distribution choice, the label probabilities have a stair-like shape with a \emph{constant} (deterministic) variation in each step.

 As detailed in the next section, our approach circumvents the need to pre-define a parametric uni-modal model for network outputs, and to set its parameters. We enforce uni-modality and a consistent order between the labels, but without constraining the learned representation to any specific parametric model, allowing the training to explore a larger space of solutions. To this end, we describe the uni-modality property through ordering adjacent labels, thereby ensuring decreasing monotonicity of the probabilities on both sides of the target label. Such an order is represented using a set of inequality constraints on all pairs of adjacent labels. The competing constraints are optimized with a powerful extension of the log-barrier method \citep{boyd2004convex,kervadec2019log}, which is well-known in the context of interior-point methods in convex optimization. Unlike \citep{beckham2017unimodal}, our method scales up to a very large number of labels. Furthermore, we provide a new metric, as a proxy, to asses the uni-modality of a distribution by measuring the order between adjacent labels.

\section{Uni-modality via pairwise inequality constraints}
\label{sec:proposed-method}

Let us consider a set of training samples ${\mathbb{D} = \{(\bm{X}^{(i)}, y^{(i)})\}_{i=1}^n}$ where ${\bm{X}^{(i)}}$ is an input sample, a realization of the discrete random variable ${\mathbf{X}}$ with support set ${\mathcal{X}}$; ${y^{(i)}}$ is the sample label, a realization of the discrete random variable ${\mathbf{y}}$ with support set ${\mathcal{Y} = \{1, \cdots, c\}}$ that exhibits an \emph{overall order} between the labels,
${y_1 \before y_2 \before \dots \before y_{c-1} \before y_{c} \; , }$
where ${a \before b}$ means that the event described by the label $a$ \emph{is ordered before} the event described by the label ${b}$. In this work, we propose to use pairwise inequality constraints to enforce implicitly the uni-modality of the posterior probability, with the latter decreasing monotonically as we move further away from the target label. For notation simplicity and clarity in this section, we omit sample index ${(\cdot)^{(i)}}$ and the expected value of losses over all the samples. We define  ${\hat{p}(\mathbf{y} | \mathbf{X})}$ as the posterior probability estimated by a neural-based model (function) ${\mathcal{M}(\bm{X};\; \bm{\theta})}$. ${\bm{s} \in \R^c}$ denotes the logit scores\footnote{For generality, we assume that such scores are unbounded.} obtained by the model ${\mathcal{M}}$, where the posterior probability is computed using standard \verb+softmax+ function, ${\hat{p}(k | \bm{X}) = \exp(\bm{s}_k) / \sum_{j=1}^c \exp(\bm{s}_j)}$. Let ${\dom{f}}$ and ${\ran{f}}$ denote the domain and range of function $f$, respectively.

We describe the uni-modality of a function with respect to a target point (or label) as a decreasing monotonicity of the function above and below the point, according to some pre-defined order. To ensure such decreasing monotonicity in a non-parametric way, we embed hard pairwise constraints on the order between every two adjacent points, within each of two sets of points, one including those below the target and the other including those above. Instead of ordering probabilities, we consider ordering scores.
For a sample ${(\bm{X}, y)}$, and its score vector ${\bm{s}}$, we formulate adjacent ordering as a constrained-optimization problem using the following set of hard pairwise inequality constraints:
\begin{equation}
\label{eq:eq-2}
\begin{aligned}
& \underset{\bm{\theta}}{\text{minimize}}
& & \mathcal{C}(\mathcal{M}(\bm{X}\;; \bm{\theta}), y) \\
& \text{subject to}
& &
\bm{s}_k < \bm{s}_{k +1}, \; \text{for } k < y \;,
\quad  \text{and} \quad   \\
&&&
\bm{s}_{k +1} < \bm{s}_k, \;  \text{for } y \leq k < c \; ,
\end{aligned}
\end{equation}
where ${\mathcal{C}(\cdot, \cdot)}$ is a standard classification loss such as the cross-entropy (CE):
$\mathcal{C}(\mathcal{M}(\bm{X};\; \bm{\theta}), y) = -\log{\hat{p}(y | \bm{X})}$.
The cross-entropy will be used in our experiments, but our constrained optimization can be integrated with any other classification loss in a straightforward manner.

Our constrained optimization problem in Eq.\eqref{eq:eq-2} is very challenging for modern deep networks involving large numbers of trainable parameters \citep{kervadec2019constrained,kervadec2019log,marquez2017imposing,pathak2015constrained,ravi2018constrained}. In the context of deep networks, hard constraints are typically addressed with basic penalty methods \citep{he2016learning,jia2017constrained,kervadec2019constrained}  as they accommodate SGD optimization, avoiding explicit primal-dual steps and projections. However, for a large set of constraints, penalty methods might have difficulty guaranteeing
constraint satisfaction as they require careful and manual tuning of the weight of each constraint. In principle, standard Lagrangian-dual optimization seeks automatically the optimal weight of the constraints, and have well-established advantages over penalty methods, in the general context of convex optimization \citep{boyd2004convex}. However, as shown and discussed in several recent deep learning works \citep{marquez2017imposing,pathak2015constrained,ravi2018constrained,kervadec2019constrained,kervadec2019log}, in problems other than ordinal classification, those advantages do not materialize in practice for deep networks due mainly to the interplay between the dual steps and SGD optimization, causing instability, and to the incurred computational complexity. To solve our problem in Eq.\eqref{eq:eq-2}, we consider two alternatives to explicit Lagrangian-dual optimization: a penalty-based method and a powerful extension of the log-barrier method, which is well-known in the context of interior-point methods \citep{boyd2004convex,kervadec2019log}. In particular, the log-barrier method is well-suited to our problem. It approximates Lagrangian optimization via implicit dual variables, handling effectively large numbers of constraints, while accommodating standard SGD and avoiding explicit Lagrangian-dual steps and projections.

\textbf{Penalty-based optimization:} ${\mathcal{C}}$ in Eq.\ref{eq:eq-2} is augmented by converting each inequality constraint into a penalty term ${\mathcal{H}}$ that increases when the corresponding constraint is violated \citep{bertsekas1995athena}. To impose the inequality constraint $a < b$, a quadratic penalty is used,
${
    \mathcal{H}(\delta_a^b) =
    (\delta_a^b + \epsilon)^2 \;  \text{if } \delta_a^b \geq 0. \;
    0  \; \text{otherwise} \; ,
}$
where ${\delta_a^b = a - b}$, ${a, b \in \R}$, and ${\epsilon \in \R_+^*}$ is a slack constant to avoid the equality case. In this case, our problem in Eq.\eqref{eq:eq-2} becomes,
\begin{align}
    \label{eq:eq-quad-pen}
    & \underset{\bm{\theta}}{\text{minimize}} \quad \mathcal{C}(\mathcal{M}(\bm{X};\; \bm{\theta}), y)
    +
    \lambda \Big(
    \sum_{j=1}^{y-1} \mathcal{H}(\Delta_k^{k+1}(\bm{s})) +
    \sum_{j=y}^{c-1} \mathcal{H}(\Delta_{k+1}^{k}(\bm{s}))\Big) \; ,
\end{align}
where ${\Delta_m^l(\bm{s}) = \bm{s}_{m} - \bm{s}_l}$, and ${\lambda \in \R_+}$ is a model hyper-parameter, which balances the contribution of all the penalties encouraging constraint satisfaction in Eq. \eqref{eq:eq-quad-pen}; and determined using a validation set. This method is referred to as PN (Eq.\ref{eq:eq-quad-pen}).

\textbf{Log-barrier optimization:}
While PN-based methods are simple and straightforward, they do not guarantee constraints satisfaction and require an empirical tuning of the importance coefficient(s) \citep{fletcher1987practical,gill1981practical}. Moreover, once a constraint is satisfied, the penalty is zero. Consequently, constraints that are satisfied in one iteration may not be satisfied in the next one since the penalty does not play a role of a barrier at the feasible set of solutions. This can be problematic when dealing with a large number of constraints at once. To avoid such well known issues with PN methods, we consider log-barrier methods (LB) as an alternative. LB-methods belong to interior-point methods (IP) \citep{boyd2004convex}, which aim to approximate Lagrangian optimization with a sequence of unconstrained problems and implicit dual variable, avoiding dual steps and projections \citep{boyd2004convex}.  Eq.\eqref{eq:eq-2} can be re-written in a standard form of a LB method:
\begin{equation}
\label{eq:eq-5}
\begin{aligned}
& \underset{\bm{\theta}}{\text{minimize}}
& & \mathcal{C}(\mathcal{M}(\bm{X}\;; \bm{\theta}), y) \\
& \text{subject to}
& & \Delta_k^{k+1}(\bm{s}) < 0 , \; \text{for } k < y ,
\quad \text{and} \quad \\
&&& \Delta_{k+1}^{k}(\bm{s}) < 0, \;  \text{for } y \leq k < c \;.
\end{aligned}
\end{equation}
LB methods are widely used for inequality constrained problems \citep{boyd2004convex}. The main aim of LB methods is to convert a constrained problem of the form in \eqref{eq:eq-5}) into an unconstrained one via an indicator function ${\mathcal{I}(\cdot)}$ that has zero penalty when the constraint is satisfied, and a penalty of ${+\infty}$ otherwise. Instead of using ${\mathcal{I}}$, LB methods employ an approximate, ${\hat{\mathcal{I}}}$, using the logarithmic function, where the penalty decreases the further we get away from violating the inequality, forming a barrier between feasible and infeasible solutions. To solve Eq.\eqref{eq:eq-5}, a strictly feasible set of parameters ${\bm{\theta}}$ are required as a starting point. Such a set is found through Lagrangian minimization of inequality constraints (\emph{phase I} \citep{boyd2004convex}), which turns out to be a problem of similar difficulty as the constrained optimization in Eq.\eqref{eq:eq-5} \citep{boyd2004convex}. Such strategy makes standard LB methods impractical for constraining deep models. We use an extension of the standard log-barrier based on a different approximation of the indicator function \citep{kervadec2019log}. The main advantage is that this algorithm does not require starting from a feasible solution -- i.e., ${\dom{\hat{\mathcal{I}}}}$ is no longer restricted to feasible points of ${\bm{\theta}}$. A direct consequence is that stochastic optimization techniques such as SGD can be directly applied without the need for a feasible starting point. We replace problem Eq.\eqref{eq:eq-5} by the following sequence of unconstrained problems, parameterized by a temperature $t$:
\begin{align}
    \label{eq:eq-elb}
    & \underset{\bm{\theta}}{\text{minimize}} \quad \mathcal{C}(\mathcal{M}(\bm{X};\; \bm{\theta}), y)
    + \Big(\sum_{j=1}^{y-1} \hat{\mathcal{I}}(\Delta_k^{k+1}(\bm{s})) + \sum_{j=y}^{c-1} \hat{\mathcal{I}}(\Delta_{k+1}^{k}(\bm{s}))\Big) \; ,
\end{align}
where ${\hat{\mathcal{I}}}$ is a log-barrier extension, which is convex, continuous, and twice differentiable \citep{kervadec2019log}:
\begin{align}
    \label{eq:eq-7}
    \hat{\mathcal{I}}(r) =
    \begin{cases}
    - \frac{1}{t} \log(-r) & \text{if } r \leq -\frac{1}{t^2} \;, \\
    tr - \frac{1}{t} \log(\frac{1}{t^2}) + \frac{1}{t} & \text{otherwise}\; .
    \end{cases}
\end{align}
One can show that optimizing log-barrier extensions approximate Lagrangian optimization of the original constrained problem with implicit dual variables, with a duality gap upper bounded by a factor of $1/t$; see Proposition 2 in \citep{kervadec2019log}.
Therefore, in practice, we use a varying parameter $t$ and optimize a sequence of losses of the form \eqref{eq:eq-elb}, increasing gradually the value of $t$ by a constant factor. The network parameters evaluated at the current $t$ and epoch are used as a starting point for the next $t$ and epoch.
In our experiments, we refer to this LB method as the \emph{extended log-barrier} (ELB) (Eq.\ref{eq:eq-elb}).

\section{Experiments}
\label{sec:experiments}

\noindent \textbf{Performance metrics:}
We denote ${\hat{y}}$ as the label predicted by the model, ${\E}$ as the expectation value, and ${\mathbb{T}}$ as an evaluation dataset. For our evaluation, we report performances using two metrics:
1) The mean absolute error, which is often used in OC setups: ${\texttt{MAE} = \E\limits_{(\bm{X}, y) \in \mathbb{T}} \big[\abs{\hat{y} - y} \big]}$; and
2) We propose a new metric that measures how well labels are ordered, above and below the true or predicted label.
Following the \emph{non-monotonic index} defined in \citep{ben1995monotonicity,gutierrez2016current}, we propose the \emph{Side-Order Index} (SOI) that counts the number of satisfactions (non-violations) of the order constraints over adjacent labels above and below a reference label $\nu$,
    ${
        \texttt{SOI}_{\nu} =
        \E_{(\bm{X}, y) \in \mathbb{T}}
        \big[ \frac{1}{c-1} (
        \sum_{j=1}^{\nu-1}  \ind_{\Delta_j^{j+1}(\hat{p}) < 0}
        + \sum_{j=\nu}^{c-1} \ind_{\Delta_{j+1}^{j}(\hat{p}) < 0}
         )
        \big] \; .
    }$
    This metric can be seen as a proxy to describe how well a distribution is uni-modal with respect to a reference label ${\nu}$.
    Furthermore, it is appropriate for evaluating the performance of a constrained-optimization method for our problem
    in \eqref{eq:eq-2} as it evaluates constraint satisfaction. We compute the expected value over a normalized measure so that the metric is independent of the total number of labels, hence, independent of the number of pairs of ${c - 1}$ adjacent labels. Consequently, the range of the measure is in ${[0, 1]}$, where $0$ indicates that all the adjacent labels are un-ordered, and $1$ indicates a perfect order.   In our experiments, we consider the case where the reference label ${\nu}$ is the predicted label. It is noted ${\texttt{SOI}_{\hat{y}}}$. In this case, we measure the consistency of the model's predictions with respect to the predicted label and \emph{independently} from the true label. This measure is the most important as it assesses the model's consistency when it is evaluated in a real scenario, where the true label is unknown.

\noindent  \textbf{Datasets and training protocol:}
We consider datasets that naturally exhibit order between the labels.  We target datasets that have large number of labels. Applying OC over datasets with very small number of labels is unlikely to show the power and limitation of different methods. We consider 3 different applications: breast cancer grading, photographs dating, and age estimation using public benchmarks.
 (1) \emph{Breast cancer grading}: This task consists in classifying histology images of breast biopsy into different grades that are ordered with respect to the cancer aggressivity. We use the dataset BACH (Part A) Breast cancer \citep{aresta2018bach} referred to here as ICIAR. The dataset contains a total of 400 images, and 4 classes: normal, benign, in Situ, and invasive (in this order). Following the protocol described in \citep{belharbi2019wsolentropy,rony2019weak-loc-histo-survey}, we perform two splits of the dataset where in each split we take 50\% of samples per class for test, and perform 5-fold cross-validation to build the train/validation sets.
 (2) \emph{Photographs dating}: This task consists of predicting in which decade a color image was taken. We consider the dataset Historical Color Image dataset (for classification by decade)\footnote{\url{http://graphics.cs.cmu.edu/projects/historicalColor/}} \citep{palermo2012datingdataset}, referred to here as HCI, along with the experimental protocol in \citep{palermo2012datingdataset}. The dataset contains 5 decades from 1930s to 1970s, each containing 265 images, for a total of 1325 images. We took 50 random images per decade to form a test set, with the rest of the data used for training and validation sets using 10-fold cross-validation. This process is repeated 10 times.
 (3) \emph{Age estimation}: This task consists of estimating someone's age based on their picture. We consider three datasets:
 (q) FG-NET dataset\footnote{\url{https://yanweifu.github.io/FG\_NET\_data/index.html}} \citep{panis2016overview}, referred to here as FGNET: It is a very early database used for age estimation, which contains 1,002 face images from 82 individuals, with ages ranging from 0 to 69 (70 classes);
 (b) Asian Face Age Dataset (AFAD)\footnote{\url{http://afad-dataset.github.io/}} \citep{niu2016ordinal} light (AFAD-Lite), with a total of 59,344 samples, and age ranges from 15 to 39, and a total number of classes of 22; and
 (c) AFAD-Full \citep{niu2016ordinal}, with a total of 165,501 samples, an age ranges from 15 to 72, and a total number of classes of 58. The same protocol is conducted over the three datasets. Following the experimental setup in \citep{chang2011ordinal,chen2013cumulative,niu2016ordinal,wang2015deeply}, we randomly select 20\% of the entire dataset for testing, and perform 5-fold cross-validation to build train and validation sets. This process is repeated 10 times. All the splits are done randomly using a deterministic code that we provide publicly along with the splits. We report the mean and standard deviation of each metric.

 For a fair comparison, all the methods use exactly the same training setup including the shuffling, the order of processing samples, and the experimental environment (device and code). We use a pre-trained Resnet18 \citep{heZRS16} as a model ${\mathcal{M}}$,  with WILDCAT \citep{durand2017wildcat} pooling layer and  hyper-parameters ${kmax=0.1, \alpha=0}$. Randomly cropped patches of size $256 \times 256$ are used for training. We train for 1000 epochs using SGD with a learning rate of ${0.001}$,\footnote{Except for HCI dataset, where we use a learning rate of ${0.0001}$.}  which is decayed every ${100}$ epochs by ${0.1}$, with an allowed minimum value of ${1e-7}$, a batch size of ${8}$, a  momentum of ${0.9}$ and weight decay of ${1e-5}$. For PN (Eq.\ref{eq:eq-quad-pen}), ${\lambda}$ is selected using validation from the set ${\{1e-1, 1e-2, 1e-3, 1e-4, 1e-5\}}$, and set to ${1e-2}$; ${\epsilon=1e-1}$. For LB methods, ${t}$ is initialized to ${1}$, and iteratively increased after each epoch by a factor of ${1.001}$, with an allowed maximum value of ${5}$. Due to the large size of AFAD-Lite-Full datasets, training is done using a batch size of 64 for 100 epochs.\footnote{For AFAD-Full, PN and ELB are trained only for 80 epochs due to time constraint.} For AFAD-Full, $t$ is initialized to $4.5$ and increased by a factor of ${1.01}$. For the case of age estimation task, we do not use any face detector, face cropping, nor face alignment. We feed the network the raw image. For efficient computation and scalability, we implement the difference ${\Delta_m^l(\bm{s})}$ (Eq.\ref{eq:eq-quad-pen}, \ref{eq:eq-5}) using 1D convolution with fixed weights ${[+1, -1]}$ as a differentiator from left to right. The difference in the other direction is the same as the one from left to right but with sign ${(-)}$.



\section{Results and discussion}
\label{sec:results}
\begin{table*}[ht!]
  \caption{Evaluation of different methods over the test sets of ICIAR and HCI datasets (classification datasets).}
  \label{tab:iciar-hci}
  \centering
  \small
  \resizebox{0.9\linewidth}{!}{
  \begin{tabular}{l|l|l||l|l}
    \hline
    Method &  \multicolumn{2}{c||}{ICIAR} &  \multicolumn{2}{c}{HCI}\\
    \cmidrule{2-5}
                      &   ${\texttt{MAE}}$  & ${\texttt{SOI}_{\hat{y}}}$ (\%)    &     ${\texttt{MAE}}$  & ${\texttt{SOI}_{\hat{y}}}$ (\%) \\
    \midrule
    CE       & ${{0.19 \pm 0.02}}$ &  ${80.05 \pm 1.68}$   &  $0.68 \pm 0.05$  & $78.05 \pm 1.67$ \\
    \hline
    REN \citep{cheng2008neural}           & $0.23 \pm 0.028$  & $54.94 \pm 1.38$  &    ${{0.63 \pm 0.03}}$ & ${57.08 \pm 1.46}$ \\
    LD \citep{geng2016label}             & ${0.49 \pm 0.03}$ &  ${91.02 \pm 1.33}$   &    ${1.00 \pm 0.04}$  & ${84.42 \pm 1.33}$ \\
    MV \citep{pan2018mean}               & ${0.61 \pm 0.04}$  & ${62.99 \pm 2.09}$   &  ${0.69 \pm 0.03}$  & ${71.82 \pm 1.93}$ \\
    PO \citep{beckham2017unimodal}        & ${0.60 \pm 0.02}$  & ${61.27 \pm 1.63}$   &  ${0.71 \pm 0.03}$  & ${49.37 \pm 1.69}$ \\
    \hline
    PN (ours)   & ${0.19 \pm 0.01}$  & ${84.69 \pm 1.63}$   &  ${0.64 \pm 0.05}$  &  ${83.06 \pm 1.44}$ \\
    ELB (ours)   & ${\bm{0.17 \pm 0.02}}$   & ${\bm{99.32 \pm 0.28}}$   &  $\bm{0.62 \pm 0.05}$ & $\bm{99.15 \pm 0.39}$ \\
    \hline
  \end{tabular}
  }
\end{table*}

\begin{table*}[ht!]
  \caption{Evaluation of different methods over the test sets of AFAD-Lite, AFAD-Full, and FGNET datasets. $--$ indicates that the method does not scale to a large number of classes.
  }
  \label{tab:afad-fgnet}
  \centering
  \small
  \resizebox{1.\linewidth}{!}{
  \begin{tabular}{l|l|l||l|l||l|l}
    \hline
    Method &  \multicolumn{2}{c||}{AFAD-Lite} & \multicolumn{2}{c||}{AFAD-Full} & \multicolumn{2}{c}{FGNET}\\
    \cmidrule{2-7}
                      &  ${\texttt{MAE}}$ & ${\texttt{SOI}_{\hat{y}}}$ (\%)
                      &  ${\texttt{MAE}}$ & ${\texttt{SOI}_{\hat{y}}}$ (\%)
                      &  ${\texttt{MAE}}$ & ${\texttt{SOI}_{\hat{y}}}$ (\%)  \\
    \midrule
    CE          & $3.69 \pm 0.06$  & $68.67 \pm 1.00$  & $3.73 \pm 0.06$ & $63.69 \pm 0.47$ & ${7.79 \pm 0.79}$   & ${54.11 \pm 1.18}$  \\
    \hline
    REN \citep{cheng2008neural}          &  ${3.00 \pm 0.01}$   & ${58.91 \pm 1.16}$  & ${\bm{3.19 \pm 0.02}}$   &   ${55.97 \pm 0.30}$ & ${4.53 \pm 0.44}$   & ${53.32 \pm 0.34}$ \\
    LD \citep{geng2016label}             &  ${5.03 \pm 0.02}$   & ${66.52 \pm 0.19}$  & ${5.15 \pm 0.028}$   &    ${61.82 \pm 0.12}$ & ${18.25 \pm 0.86}$   & ${51.60 \pm 0.20}$  \\
    MV \citep{pan2018mean}               &  ${\bm{2.96 \pm 0.02}}$   & ${83.32 \pm 0.95}$  & ${3.20 \pm 0.02}$   &    ${70.56 \pm 0.58}$ & ${5.21 \pm 0.65}$  & ${59.72 \pm 1.95}$ \\
    PO \citep{beckham2017unimodal}       &  ${3.56 \pm 0.02}$   & ${20.34 \pm 0.41}$  & $--$   &  $--$ & $--$  & $--$ \\
     \hline
     PN (ours) & $3.52 \pm 0.02$ & $79.72 \pm 2.20$  & $3.69 \pm 0.04$ & $78.54 \pm 1.72$ & ${5.92 \pm 0.46}$   & ${60.58 \pm 3.39}$ \\
     ELB (ours) & $3.15 \pm 0.02$ & $\bm{98.67 \pm 1.17}$  & $3.40 \pm 0.05$ & $\bm{97.02 \pm 0.88}$ & $\bm{4.27 \pm 0.47}$   & ${\bm{98.96 \pm 0.83}}$   \\
    \hline
  \end{tabular}
  }
\end{table*}

\begin{figure*}[ht!]
  \centering
  \includegraphics[width=.45\linewidth]{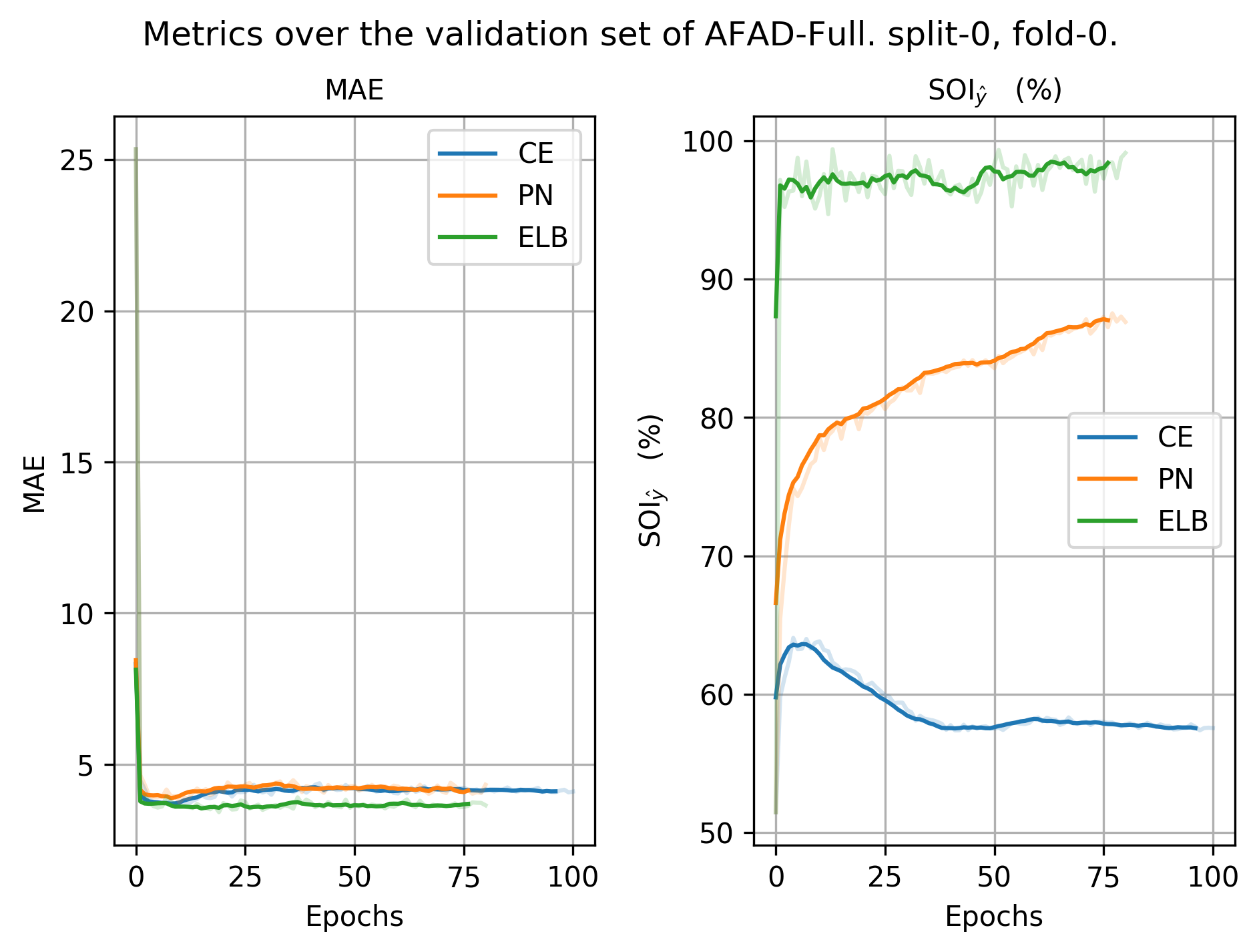}
  \includegraphics[width=.45\linewidth]{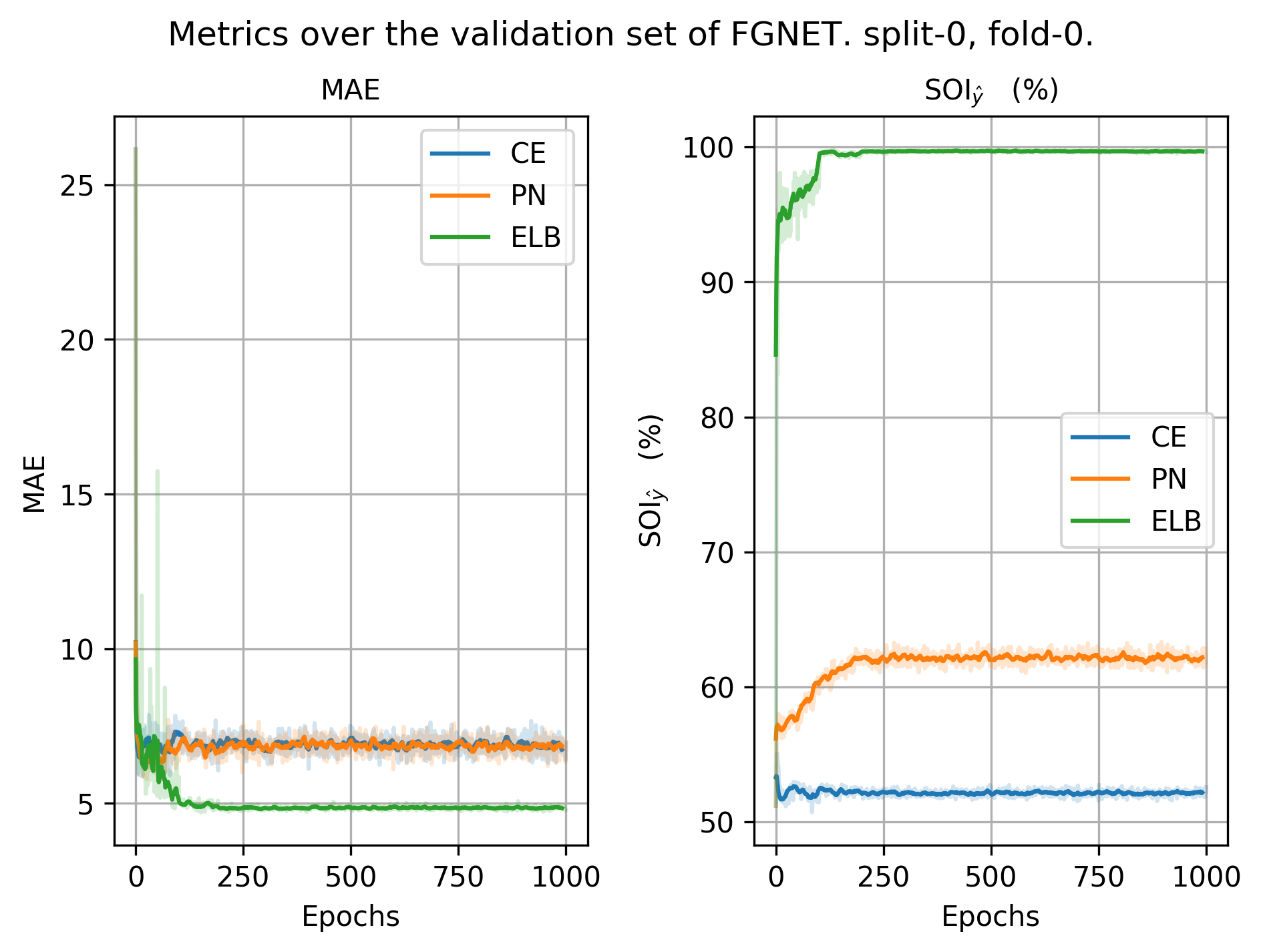}
  \includegraphics[width=.45\linewidth]{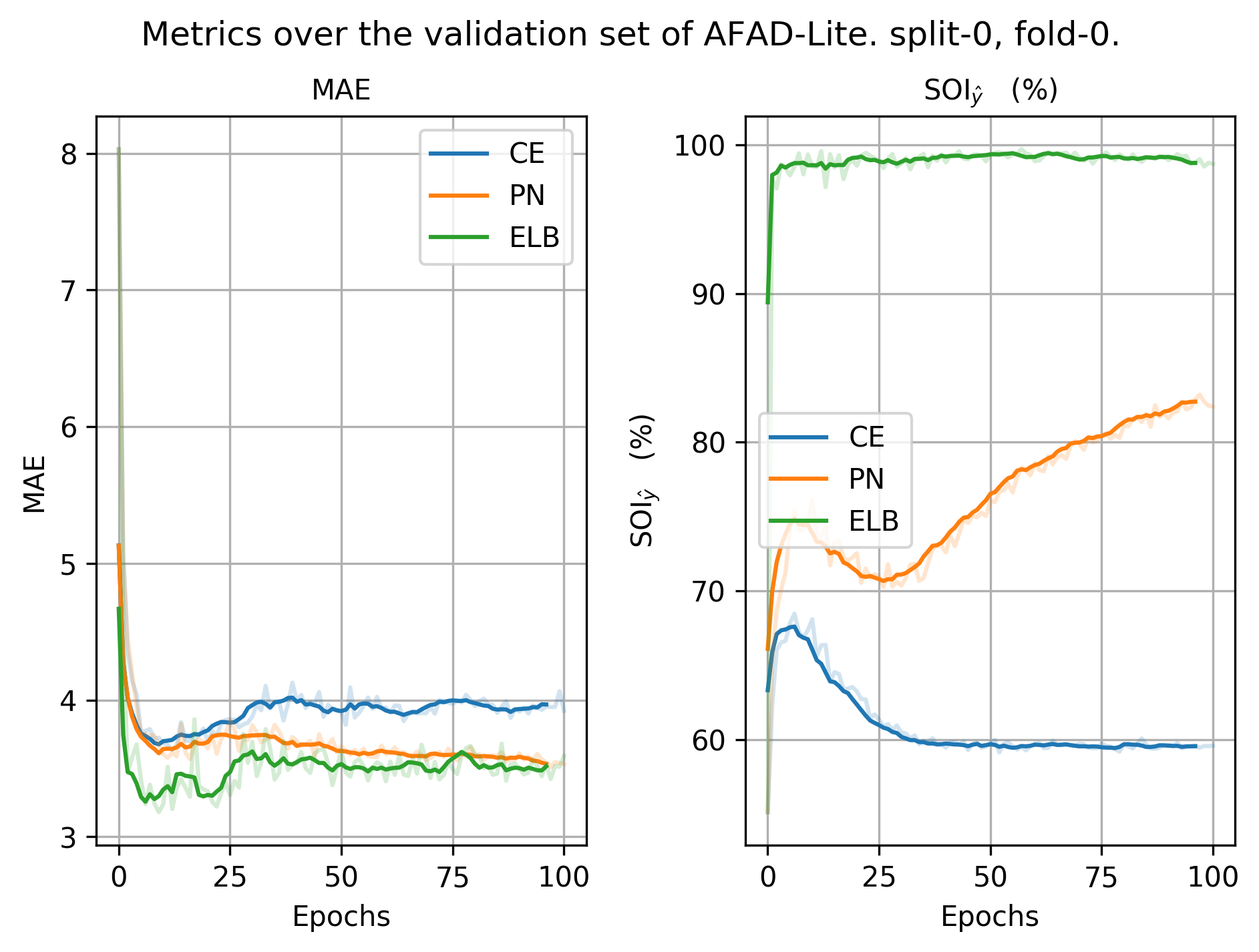}
  \includegraphics[width=.45\linewidth]{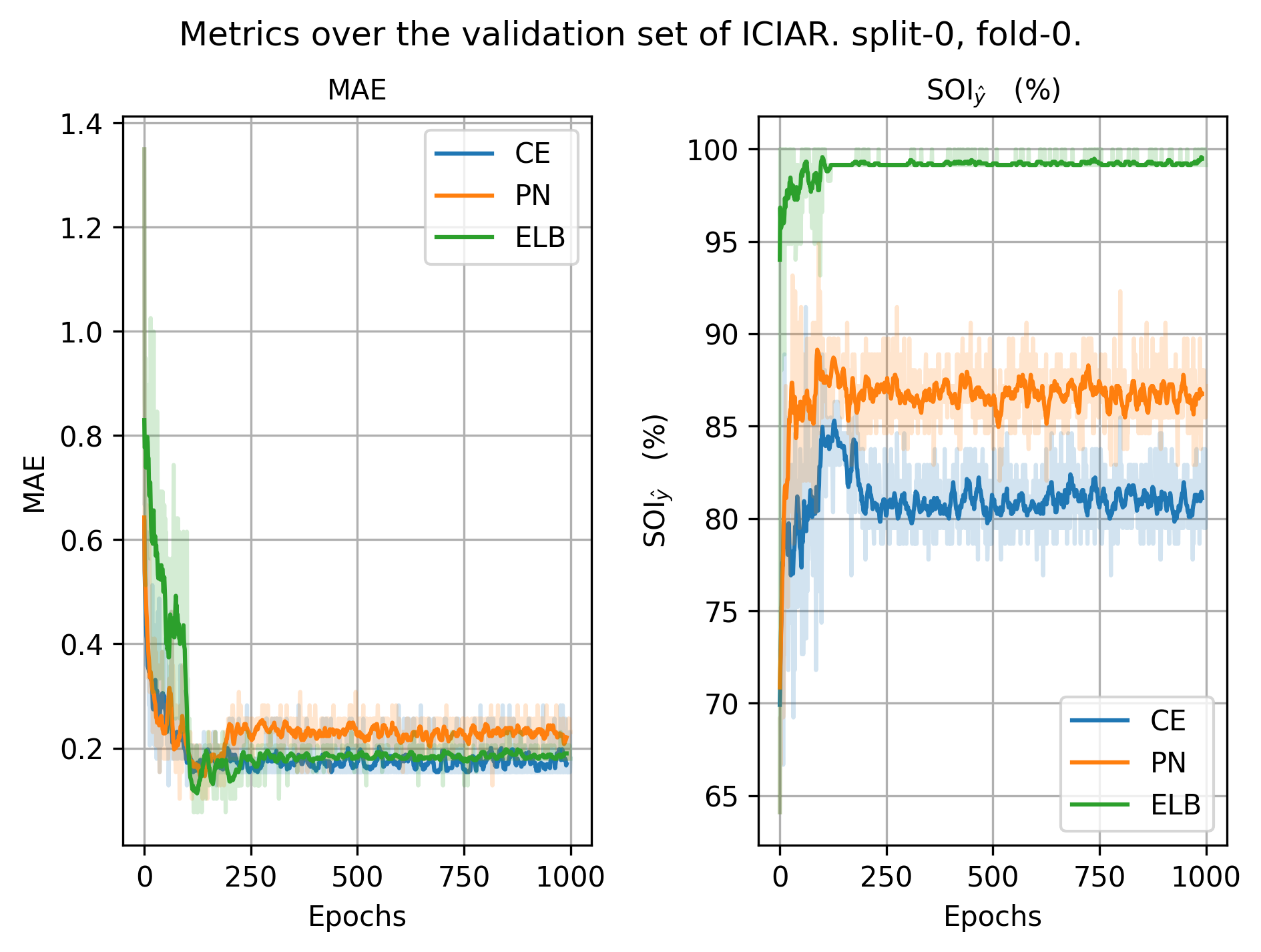}
  \caption{Moving average of ${\texttt{MAE}}$ and $\texttt{SOI}_{\hat{y}}$ metrics over the \textbf{validation set (one run over fold-0, split-0)} of AFAD-Full (top-left), FGNET (top-right), AFAD-Lite (bottom-left), and ICIAR (bottom-right) for CE, PN and ELB. (Best visualized in color.)}
  \label{fig:curves-valid-others-main}
\end{figure*}

The quantitative results obtained with the different methods over the different test sets are presented in Tabs.\ref{tab:iciar-hci}- Tab.\ref{tab:afad-fgnet}. We recall the notation of the different methods --
 CE:  cross-entropy method;
 REN \citep{cheng2008neural}: re-encode the hard target into a vector of binary values and use the mean squared error as a loss. The threshold is set to ${0.5}$;
 LD \citep{gao2017deep,geng2016label,geng2013facial,huo2016deep}: label distribution learning with Bayes rule prediction. The variance is set to ${1}$;
 MV \citep{pan2018mean}: mean-variance loss combined with softmax where the predicted label is the round function of the expected value of the labels. Following \citep{pan2018mean}, we set $\lambda_1=0.2, \lambda_2=0.05$;
 PO \citep{beckham2017unimodal}: Hard-wire Poisson distribution at the network output, and use cross-entropy for learning. ${\tau}$ is fixed and set to 1 as in the paper \citep{beckham2017unimodal}, and the prediction is based on the expected value of the labels;
 PN (Eq.\ref{eq:eq-quad-pen}): penalty-based method;
 ELB (Eq.\ref{eq:eq-elb}): extended log-barrier method.
 All the methods have the same exact model capacity (i.e., ResNet18 \citep{heZRS16}), except PO \citep{beckham2017unimodal} where we add a dense layer that maps from $c$ (number of classes) into one (positive score). Based on the obtained results, we note the following.
 \textbf{(1) In term of} ${\texttt{SOI}_{\hat{y}}}$ \textbf{consistency}, we observe that ELB method yields the best results, achieving almost perfect ordering ${\texttt{SOI}_{\hat{y}} > 97\%}$ over all datasets.
 Then, comes second the PN method, but with a large gap of ${60\% < \texttt{SOI}_{\hat{y}} < 84\%}$ in comparison to ELB, and a slightly  better performance than CE (${54\% < \texttt{SOI}_{\hat{y}} < 80\%}$). We note that the gap between PN and ELB increases with the number of classes: ${\sim15\%, \sim16\%}$ on ICIAR and HCI with $4$ and $5$ classes, respectively; ${\sim19\%, \sim19\%}$ on AFAD-Lite with ${22}$ classes, and AFAD-Full with ${58}$ classes, respectively; and ${\sim38\%}$ on FGNET with ${70}$ classes. This is expected, since the PN method does not cope well with the interplay between different constraints, more so when the number of constraints is large unlike LB methods which approximate Lagrangian optimization. \hspace{2mm} The different methods REN, LD, MV, and PO yield a ${\texttt{SOI}_{\hat{y}}}$ not far from the CE method. LD achieves good results over ICIAR and HCI with $\texttt{SOI}_{\hat{y}}$ of ${\sim91\%}$ and ${\sim84\%}$ ranking in the second place after ELB. However, its performance drops to ${\sim66\%, \sim61\%}$ over AFAD-Lite -Full, and to ${\sim51\%}$ over FGNET suggesting that it does not handle well large number of classes. Compared to LD, MV performs better on large number of classes achieving ${\sim83\%, \sim70\%, \sim59\%}$ over AFAD-Lite -Full, and FGNET, respectively. REN performance is usually worse than CE. The case of PO is particular. It is expected to obtain ${\texttt{SOI}_{\hat{y}} = 100\%}$ but since the predicted label is the expected value of the label (and not the ${\argmax}$ of the scores), it achieves a low level of performance. When PO uses ${\argmax}$, it does not represent a fair comparison to other methods since the order in PO is \emph{hardwired}, while the order is \emph{learned} with all other methods. PO \citep{beckham2017unimodal} does not scale up to large numbers of classes, and, in \citep{beckham2017unimodal}, it was evaluated only on two datasets, with 5 and 8 classes. \hspace{2mm}  These results show the amount of disorder with a CE prediction in an OC setup, which confirms its inadequacy to such context. Furthermore, we observe the power of LB methods compared to PN method, with the former achieving much better constraint satisfaction than the latter. This shows the effectiveness of LB methods and, in particular, its extension \citep{kervadec2019log}, for optimizing with inequality constraints
 on a model output (Fig.\ref{fig:curves-valid-others-main}).
 \textbf{(2) In term of} ${\texttt{MAE}}$, in all the datasets, we observe that combining CE with inequality constraints always helps improving the performance. Over ICIAR and HCI datasets, ELB obtained the best performance with ${\texttt{MAE} = 0.17, 0.62}$, respectively. Over AFAD-Lite, the MV method yields ${\texttt{MAE} = 2.96}$, while REN obtains a state-of-the-art error over AFAD-Full with ${\texttt{MAE} = 3.19}$, compared to  ${\texttt{MAE}=3.34}$ reported in \citep{niu2016ordinal}. In both datasets, ELB ranks third with ${\texttt{MAE} = 3.15, 3.40}$, respectively. All our experiments are repeated 10 times, while \citep{niu2016ordinal} repeats experiments for 100 times.
 On the FGNET dataset, which has 70 labels, ELB obtained the best performance of ${\texttt{MAE} = 4.27}$,  followed by REN, and MV with ${\texttt{MAE} = 4.53, 5.21}$, respectively. Combining the inequality constraints with the CE to promote a consistent output prediction helps always to improve the ${\texttt{MAE}}$ performance.
 \textbf{(3) Training time}: PN and LB methods do not add a significant computation overhead compared to CE. \textbf{(4) Which method to choose in practice?}: this is an important question and the answer depends mostly on the specific application.  Based on the above empirical evidence we suggest that: (\emph{a}) for critical applications where an agent is used who expects an explanation for the model’s decision (e.g., in the medical domain, where model interpretability/consistency is a priority),  our method is a better choice; (\emph{b}) For applications where the ${\texttt{MAE}}$ performance is a priority (e.g., control and automatic applications) without agent, other methods can be a good choice. However, our method can be considered as well since it yields competitive ${\texttt{MAE}}$. (\emph{c}) for applications where both metrics are crucial, our method is a reasonable choice. \hspace{2mm} Independently from the chosen method, the proposed ${\texttt{SOI}}$ metric provides a helpful tool for the agent to quickly and reliably assess the prediction’s consistency along with the PD visualization.


\section{Conclusion}
\label{sec:conclusion}

We presented a new constrained-optimization formulation for ordinal classification, with uni-modality of the label distribution imposed implicitly via a set of inequality constraints over pairs of adjacent labels.
To tackle the ensuing challenging optimization problem, we solve a sequence of unconstrained losses based on a powerful extension of the log-barrier method, which is well-known in the context of interior-point methods. This accommodates standard SGD, and avoids computationally expensive Lagrangian dual steps and projections, while outperforming substantially standard penalty methods. Our non-parametric model is more flexible than the existing ordinal classification techniques: it does not restrict the learned representation to a specific parametric model, allowing the training to explore larger spaces of solutions and removing the need for {\em ad hoc} choices, while scaling up to large numbers of labels. It can be used in conjunction with any standard classification loss and any deep architecture. We also propose a new performance metric for ordinal classification, as a proxy to measure a distribution uni-modality, referred to as the Sides Order Index (SOI). We report comprehensive evaluations and comparisons to state-of-the-art methods on benchmark public datasets for several ordinal classification tasks, showing the merits of our approach in terms of label consistency and scalability.

\medskip

\small

\bibliographystyle{apalike}
\bibliography{bibliography}

\end{document}